\documentclass{article}

\usepackage{PRIMEarxiv}

\usepackage[utf8]{inputenc} % allow utf-8 input
\usepackage[T1]{fontenc}    % use 8-bit T1 fonts
\usepackage{hyperref}       % hyperlinks
\usepackage{url}            % simple URL typesetting
\usepackage{booktabs}       % professional-quality tables
\usepackage{amsfonts}       % blackboard math symbols
\usepackage{nicefrac}       % compact symbols for 1/2, etc.
\usepackage{microtype}      % microtypography
\usepackage{lipsum}
\usepackage{fancyhdr}       % header
\usepackage{graphicx}       % graphics
\graphicspath{{media/}}     % organize your images and other figures under media/ folder

\title{HGT-Scheduler: Deep Reinforcement Learning for the Job Shop Scheduling Problem via Heterogeneous Graph Transformers
}

\author{
  Bulent Soykan \\
  Institute for Simulation and Training \\
  University of Central Florida \\
  Orlando, FL, USA\\
  \texttt{Bulent.Soykan@ucf.edu} \\
}

\begin{document}
\maketitle

\begin{abstract}

The Job Shop Scheduling Problem (JSSP) is commonly formulated as a disjunctive graph in which nodes represent operations and edges encode technological precedence constraints as well as machine-sharing conflicts. Most existing reinforcement learning approaches model this graph as homogeneous, merging job-precedence and machine-contention edges into a single relation type. Such a simplification overlooks the intrinsic heterogeneity of the problem structure and may lead to the loss of critical relational information. To address this limitation, we propose the Heterogeneous Graph Transformer (HGT)-Scheduler, a reinforcement learning framework that models the JSSP as a heterogeneous graph. The proposed architecture leverages a Heterogeneous Graph Transformer to capture type-specific relational patterns through edge-type-dependent attention mechanisms applied to precedence and contention relations. The scheduling policy is trained using Proximal Policy Optimization. The effectiveness of the proposed method is evaluated on the Fisher--Thompson benchmark instances. On the FT06 instance, the HGT-Scheduler achieves an optimality gap of 8.4\%, statistically outperforming both an identical architecture that ignores edge types ($p = 0.011$) and a standard Graph Isomorphism Network baseline. On the larger FT10 instance, the approach demonstrates favorable scalability. However, under a 50,000-step training limit, the performance of heterogeneous and homogeneous graph models is comparable, suggesting that edge-type awareness requires longer training horizons for larger problem instances. Ablation analyses further indicate that a three-layer attention architecture provides the best performance. Overall, the results confirm that explicitly modeling distinct edge semantics improves the learning of effective scheduling policies. Source code and data are available at: \href{https://github.com/bulentsoykan/HeterogeneousGraphTransformer4JSSP}{github.com/bulentsoykan/HeterogeneousGraphTransformer4JSSP}.

\end{abstract}

% keywords can be removed

\section{Introduction}

% The Job Shop Scheduling Problem (JSSP) is a fundamental problem in operations research. It models the allocation of shared resources over time. A set of jobs must be processed on a set of machines. Each job consists of a strict sequence of operations. Each operation requires a specific machine for a fixed amount of time. An operation cannot start until its preceding operation in the same job is complete. Furthermore, a machine can only process one operation at a time. The typical objective is to minimize the makespan. The makespan is the total time required to complete all jobs. JSSP is a strongly $\mathcal{NP}$-hard problem \cite{garey1976complexity,pinedo_scheduling}. As the number of jobs and machines grows, the combinatorial search space expands exponentially.

% .......................................................
The Job Shop Scheduling Problem (JSSP) is a classical and highly challenging combinatorial optimization problem in operations research and manufacturing systems. It involves sequencing a set of jobs, each consisting of a sequence of operations, on a set of machines such that all precedence and resource constraints are satisfied while minimizing objectives such as makespan or total completion time. From a computational complexity perspective, the JSSP is known to be \textit{$\mathcal{NP}$-hard} \cite{garey1976complexity,pinedo_scheduling}, and in its most general form it is commonly described as \textit{strongly $\mathcal{NP}$-hard}, indicating that no polynomial-time algorithm can solve all instances optimally \cite{Smit2025,Intelli2025}. This inherent complexity arises from the exponential growth in the number of feasible schedules as the numbers of jobs and machines increase, leading to a combinatorial explosion in the solution space.

Exact optimization methods, including mixed-integer programming and branch-and-bound algorithms, can solve only small problem instances within reasonable computational time, but they become computationally infeasible for larger sizes due to exponential time complexity \cite{Gromicho2012,Zupan2024}. As a consequence, even moderately sized instances (e.g., 10--15 jobs on 10--15 machines) remain intractable for exact solvers without extensive computation \cite{Gromicho2012}. The significant computational burden of exact optimization has motivated the development of heuristic, metaheuristic, and machine learning–based approaches that aim to produce high-quality near-optimal solutions within practical time budgets \cite{Smit2025,Learn2025}. Recent research further highlights that heuristic and learning-based methods are increasingly adopted to address large-scale JSSPs where traditional exact optimization fails due to unreasonable computational expense \cite{Wang2023}.

% .......................................................
% ////////////////////////////////////////////////////////////////////////////////////
% Traditional approaches to solving the JSSP fall into two broad categories. Exact methods guarantee optimal solutions. These include branch-and-bound algorithms and mixed-integer linear programming \cite{applegate1991computational}. However, exact methods are computationally intractable for large problem instances. They often fail to find solutions within practical time limits. Because of this, industrial practitioners rely heavily on heuristic rules. Common examples include the Shortest Processing Time (SPT) and Longest Processing Time (LPT) rules. These priority dispatching rules are fast. They compute scheduling decisions in real time. Yet, they are inherently short-sighted. They evaluate decisions using only local information. They ignore the global state of the shop floor. This lack of global awareness often leads to poor overall schedule quality \cite{panwalkar1977survey}.
% ///////////////////////////////////////////////////////////////////////////////////////////

% ***************************************************************************************
Traditional approaches to solving JSSP can generally be classified into two broad categories: exact optimization methods and heuristic approaches. Exact methods guarantee optimal solutions and include techniques such as branch-and-bound algorithms and mixed-integer linear programming \cite{applegate1991computational}. While these methods are theoretically appealing, their applicability is limited in practice due to the combinatorial explosion of the solution space. As the number of jobs and machines increases, the computational effort required by exact algorithms grows exponentially, rendering them impractical for large-scale scheduling problems.Consequently, industrial practitioners frequently rely on heuristic scheduling strategies that provide fast and reasonably good solutions within limited computational time. Among the most widely used approaches are priority dispatching rules such as the Shortest Processing Time (SPT) and Longest Processing Time (LPT). These rules are computationally efficient and enable real-time scheduling decisions in dynamic manufacturing environments. However, dispatching rules typically rely on local information when selecting the next operation to process. As a result, they often ignore the global state of the production system, which can lead to suboptimal scheduling performance \cite{panwalkar1977survey}. To address the limitations of both exact optimization and simple heuristics, recent research has increasingly focused on advanced optimization and learning-based approaches. More recently, machine learning and reinforcement learning techniques have emerged as promising tools for solving complex scheduling problems by learning decision policies directly from data and system interactions \cite{zhang2020learning,park2021learning}. These approaches aim to balance computational efficiency with solution quality, making them particularly attractive for large-scale and dynamic manufacturing environments.
Deep reinforcement learning (DRL) has recently emerged as a promising alternative for solving complex scheduling problems. In DRL-based formulations, the scheduling process is modeled as a Markov Decision Process (MDP) \cite{sutton2018reinforcement,li2017deep}. Within this framework, an agent learns a scheduling policy through repeated interaction with the environment. At each decision step, the agent observes the current state of the shop floor and selects an eligible operation to schedule next. A reward signal reflecting the quality of the resulting schedule is then provided to guide the learning process. Through exploration of different action sequences and iterative policy updates, the agent gradually learns strategies that maximize long-term cumulative rewards. Once trained, a DRL-based scheduler can generate decisions extremely quickly, behaving similarly to a fast heuristic. However, unlike traditional dispatching rules, a well-trained DRL agent can capture complex global patterns within the scheduling environment and learn to balance competing priorities across the entire shop floor \cite{wang2022deep}.

To apply DRL effectively, it is essential to represent the state of the JSSP in a form that can be processed by neural networks. The disjunctive graph is the standard mathematical representation of the JSSP \cite{balas1969machine}. In this representation, nodes correspond to operations, while edges encode constraints between operations. Two distinct types of edges are typically defined. Conjunctive edges represent technological precedence constraints that enforce the fixed processing order of operations within each job. These edges are directed. In contrast, disjunctive edges capture resource constraints arising from machine sharing, indicating that multiple operations require the same machine. These edges are typically modeled as undirected until a processing order is determined. Together, these two edge types fully describe the structural and resource constraints of the scheduling environment.

Recent DRL approaches increasingly leverage graph neural networks (GNNs) to process such structured representations. By modeling operations and their interactions as graph elements, GNNs are able to capture the relational dependencies inherent in scheduling problems. A notable example is the Learning to Dispatch (L2D) framework \cite{zhang2020learning}, which employs Graph Isomorphism Networks (GIN) to learn expressive node embeddings from the disjunctive graph \cite{xu2018powerful}. Subsequent studies have further demonstrated that graph-based DRL approaches can outperform classical dispatching rules and several traditional heuristics on benchmark scheduling instances \cite{song2023flexible,tassel2021reinforcement}. Despite these promising results, many existing methods share an important limitation. Most GNN-based DRL models treat the disjunctive graph as a homogeneous structure, merging precedence edges and machine-contention edges into a single relation type. This simplification ignores the fundamentally different semantics of technological and resource constraints and may lead to the loss of important structural information within the scheduling environment \cite{liu2024graph}.
Treating the disjunctive graph as a homogeneous structure introduces a fundamental modeling limitation. In such formulations, neural networks propagate identical messages across fundamentally different types of constraints. However, the semantics of these constraints differ significantly. A message transmitted along a job-precedence edge conveys information about upstream delays and downstream processing requirements within a job sequence. In contrast, a message propagated along a machine-sharing edge reflects resource contention and waiting times among competing operations. Collapsing these heterogeneous relations into a single edge type therefore discards important structural information and forces the neural network to infer the distinction implicitly from node features alone \cite{hu2020heterogeneous}. Prior studies in graph representation learning have shown that explicitly modeling heterogeneous relations can substantially improve the expressive power of graph neural networks when dealing with complex relational data \cite{velivckovic2018graph,wang2020heterogeneous}.

Motivated by this observation, we address this structural limitation directly. Our central hypothesis is that explicitly distinguishing edge semantics within the disjunctive graph can improve scheduling policy learning. A homogeneous graph representation restricts the representational capacity of the learning agent by masking the underlying constraint structure. In contrast, a heterogeneous graph formulation preserves the inherent distinction between precedence constraints and resource contention relations. By enabling relation-specific message passing and attention mechanisms, the model can learn different interaction patterns for each constraint type, which may lead to improved scheduling decisions \cite{schlichtkrull2018modeling}.

Based on this idea, we propose the \textit{HGT-Scheduler}, a reinforcement learning agent specifically designed for the JSSP. The proposed framework represents the scheduling environment as a heterogeneous graph in which operations correspond to nodes and constraint relations correspond to typed edges. The architecture employs a Heterogeneous Graph Transformer (HGT) to process this representation. In the proposed model, separate attention weight matrices are assigned to \texttt{precedes} edges representing technological precedence constraints and \texttt{competes} edges representing machine-sharing conflicts. This design enables the model to capture relation-specific interactions while constructing rich and context-aware state embeddings for scheduling decisions.

The learned representations are used to guide action selection within a reinforcement learning framework. Specifically, the scheduling policy is trained using Proximal Policy Optimization (PPO) \cite{schulman2017proximal}, a widely used actor--critic algorithm known for stable and efficient policy updates. PPO balances exploration and exploitation while maintaining controlled policy updates through clipped objective functions, making it well suited for complex sequential decision-making problems such as job shop scheduling.
Our main contributions are summarized as follows:

\begin{itemize}

\item \textbf{Heterogeneous graph formulation for JSSP.} 
We formulate the JSSP state as a heterogeneous graph representation within a reinforcement learning framework. 
This formulation explicitly preserves the semantic distinction between precedence constraints and machine-contention constraints, allowing the learning agent to process structurally different scheduling relationships through relation-specific message passing.

\item \textbf{HGT-based reinforcement learning scheduler.} 
We propose the \textit{HGT-Scheduler}, a reinforcement learning agent based on a Heterogeneous Graph Transformer architecture. 
The model applies type-dependent attention mechanisms to \texttt{precedes} and \texttt{competes} edges, enabling the policy network to learn constraint-specific interaction patterns within the scheduling environment.

\item \textbf{Empirical evaluation on benchmark instances.} 
We evaluate the proposed approach on the classical Fisher--Thompson benchmark instances. 
On the FT06 instance, the HGT-Scheduler achieves an optimality gap of 8.4\% and statistically outperforms an otherwise identical homogeneous architecture ($p=0.011$), as well as a standard Graph Isomorphism Network (GIN) baseline. 
On the larger FT10 instance, the method demonstrates reliable scalability. 
However, heterogeneous and homogeneous representations perform similarly under a strict 50,000-step training limit, suggesting that the benefits of edge-type awareness require longer training horizons on larger problem sizes.

\item \textbf{Architectural analysis through ablation studies.} 
We conduct an ablation analysis to investigate the effect of model depth on scheduling performance. 
The results indicate that a three-layer attention architecture provides the most effective balance between representational capacity and training stability for capturing type-dependent interactions.

\end{itemize}

The remainder of this paper is organized as follows. Section~2 provides background on the JSSP and reviews related work. Section~3 presents the proposed methodology, including the heterogeneous graph formulation, the transformer architecture, and the reinforcement learning framework. Section~4 describes the experimental setup, including the benchmark instances, baseline methods, and training parameters. Section~5 reports the experimental results, including performance comparisons, statistical analyses, and convergence behavior. Section~6 presents the ablation studies evaluating network depth and model complexity. Finally, Section~7 concludes the paper and outlines directions for future research.
% *********************************************************************************************

\section{Background and Related Work}

\subsection{The Job Shop Scheduling Problem (JSSP)}
The JSSP is one of the most extensively studied combinatorial optimization problems in operations research and production systems. It models the allocation of limited manufacturing resources over time in order to process a set of jobs through a collection of machines while satisfying technological and resource constraints. The problem involves a set of $n$ jobs, denoted as $J = \{J_1, J_2, \dots, J_n\}$, that must be processed on a set of $m$ distinct machines, denoted as $M = \{M_1, M_2, \dots, M_m\}$. Each job $J_i$ consists of a strict sequence of operations that must be executed in a predetermined order. The $j$-th operation of job $i$ is denoted by $O_{ij}$. Every operation has two predefined requirements. First, it requires processing on a specific machine $M_k \in M$. Second, it requires a fixed and uninterrupted processing time $p_{ij}$ \cite{pinedo2016scheduling,shao2013hybrid}. These technological requirements define the production routing of each job within the manufacturing system.

The JSSP is governed by two primary sets of constraints. The first is the precedence constraint, which enforces the technological order of operations within a job. Specifically, operation $O_{i(j+1)}$ cannot begin until operation $O_{ij}$ has been completed. The second is the resource constraint, which arises from the limited availability of machines. Each machine can process only one operation at a time. Furthermore, once an operation begins processing on a machine, it cannot be interrupted until completion, a condition commonly referred to as the non-preemption constraint \cite{zhang2022deep}.

A schedule specifies the starting time $S_{ij}$ for every operation $O_{ij}$ in the system. A schedule is considered feasible if it satisfies all precedence and resource constraints. The completion time of an operation is defined as $C_{ij} = S_{ij} + p_{ij}$, while the completion time of a job $C_i$ corresponds to the completion time of its final operation \cite{zhang2020learning,zhang2022deep}. 

The most widely studied objective in the classical JSSP is the minimization of the makespan, which represents the maximum completion time among all jobs in the system. The makespan is denoted as $C_{max}$ and can be expressed mathematically as
\[
C_{max} = \max_{1 \leq i \leq n} (C_i).
\]
Minimizing the makespan aims to reduce the total production time required to complete all jobs, thereby improving system throughput and resource utilization \cite{pinedo2016scheduling,shao2013hybrid}.

From a computational perspective, the JSSP is known to be strongly $\mathcal{NP}$-hard \cite{garey1976complexity}. The complexity of the problem grows combinatorially with the number of jobs and machines, leading to an exponential expansion of the solution search space. As a consequence, exact optimization techniques such as branch-and-bound or mixed-integer programming become computationally impractical for large-scale instances \cite{zhang2022deep,li2022survey}. Numerous studies have demonstrated that even moderately sized instances can require prohibitive computational effort to solve optimally. Therefore, practical scheduling systems often rely on heuristic, metaheuristic, or learning-based approaches to obtain high-quality solutions within acceptable computational times \cite{zhang2022learningAIS,chen2023intelligentAIS}.

Beyond its theoretical importance, the JSSP plays a critical role in a wide range of real-world industrial applications. In semiconductor manufacturing, job shop scheduling is used to coordinate complex wafer fabrication processes involving hundreds of sequential operations and highly constrained processing equipment \cite{lee2021deep}. In flexible manufacturing systems, scheduling methods are applied to coordinate multi-machine production lines and robotic material handling systems \cite{zhang2019flexible}. Similar scheduling challenges arise in aerospace component manufacturing, where complex machining operations must be coordinated across specialized equipment \cite{jiang2020intelligent}. In addition, modern smart factories and Industry 4.0 environments increasingly rely on intelligent scheduling algorithms and machine learning–based optimization techniques to dynamically adapt production schedules in response to changing demand, machine failures, and real-time operational data \cite{wang2021industry}. These practical applications further highlight the importance of developing scalable and intelligent scheduling approaches capable of addressing the inherent complexity of the JSSP.
% ***************************************************************************

\subsection{Disjunctive Graph Representation}
The disjunctive graph is the standard mathematical representation of the JSSP. Originally introduced by Roy and Sussmann, it provides a compact and expressive formulation that captures both precedence relations and machine resource conflicts in scheduling systems \cite{roy1964problemes}. Due to its ability to represent complex scheduling dependencies, the disjunctive graph formulation has been widely adopted in both classical optimization approaches and modern learning-based scheduling methods \cite{zhang2022deep}. Recent research in intelligent manufacturing and learning-based optimization further leverages graph-based representations of scheduling problems, where disjunctive graphs serve as structured inputs to graph neural networks and reinforcement learning frameworks designed to learn scheduling policies directly from production states \cite{wang2021drlscheduling,liu2022graphlearning}.

Formally, a disjunctive graph is defined as $G = (V, C, D)$. The set $V$ represents the vertices of the graph, where each vertex corresponds to an operation $O_{ij}$ in the scheduling problem. In addition, the vertex set contains two dummy nodes: a source node $S$ and a sink node $T$. The source node represents the starting point of the scheduling process, while the sink node represents the completion of all jobs. Both dummy nodes have zero processing time and are included to simplify the representation of global start and completion conditions \cite{pinedo2016scheduling,zhang2022deep}.

The set $C$ represents the conjunctive edges, which are directed edges enforcing job precedence constraints. For each job $J_i$, a conjunctive edge connects operation $O_{ij}$ to its immediate successor $O_{i(j+1)}$, ensuring that the processing order within each job is preserved. The weight of a conjunctive edge corresponds to the processing time of the originating operation. Additional conjunctive edges connect the source node $S$ to the first operation of each job and connect the final operation of each job to the sink node $T$. As a result, the set $C$ forms a collection of disjoint directed paths from $S$ to $T$, each representing the technological sequence of operations within a job \cite{roy1964problemes,pinedo2016scheduling}.

The set $D$ represents the disjunctive edges, which encode machine resource constraints. These edges are undirected and connect operations that require the same machine. For each machine, all operations assigned to that machine form a clique in the disjunctive graph. A disjunctive edge indicates a potential processing conflict between two operations competing for the same resource, but it does not specify the processing order between them \cite{zhang2020learning,park2021learning}. In modern intelligent scheduling frameworks, these machine-conflict edges are particularly important because they explicitly encode the decision space explored by learning-based schedulers and graph-based optimization algorithms \cite{liu2022graphlearning}.

Scheduling the JSSP can therefore be interpreted as determining an orientation for every disjunctive edge in $D$. When a scheduling decision specifies that operation $A$ must precede operation $B$ on a shared machine, the corresponding disjunctive edge is directed from $A$ to $B$. A complete schedule is obtained once all disjunctive edges have been assigned a direction \cite{zhang2022deep,wang2023graph}. Learning-based approaches increasingly model this decision process as a sequential control problem in which intelligent agents iteratively orient disjunctive edges to construct feasible schedules \cite{wang2021drlscheduling}.

A directed disjunctive graph represents a feasible schedule if and only if it contains no directed cycles. The presence of a cycle would imply an infeasible logical dependency in which an operation indirectly precedes itself. In a fully directed acyclic disjunctive graph, the makespan is equal to the length of the longest path from the source node $S$ to the sink node $T$. This longest path is commonly referred to as the \emph{critical path}. Since the JSSP objective is to minimize the makespan, scheduling on a disjunctive graph can be interpreted as a minimax optimization problem in which the goal is to direct the disjunctive edges so that the length of the critical path is minimized \cite{zhang2020learning,zhang2022deep,wang2023graph}. Graph-based learning methods have recently demonstrated that incorporating the structural information contained in disjunctive graphs significantly improves the scalability and generalization capability of intelligent scheduling systems in complex production environments \cite{liu2022graphlearning}.
% *****************************************************************************

\subsection{Graph Neural Networks in Scheduling}
DRL has recently emerged as a powerful paradigm for solving complex combinatorial optimization problems, particularly those characterized by large search spaces and sequential decision structures \cite{bengio2021machine}. In this framework, the JSSP can be formulated as a sequential decision-making problem in which scheduling decisions are made incrementally. At each decision step, an agent observes the current state of the shop floor and selects an action corresponding to the next operation to schedule. The environment then transitions to a new state and returns a reward that reflects the quality of the resulting schedule, such as improvements in makespan or machine utilization \cite{zhang2020learning,zhang2022deep}. Through repeated interaction with the environment, the agent learns a policy that maximizes cumulative reward over time. Recent research in intelligent manufacturing systems has demonstrated that DRL-based scheduling policies can effectively adapt to dynamic production environments and complex operational constraints \cite{wang2021aischeduling}.

A critical challenge in applying DRL to scheduling lies in the representation of the environment state. The state of a manufacturing system is inherently structured and relational, involving complex dependencies between jobs, operations, and machines. Traditional neural architectures such as feed-forward networks require fixed-size input representations and therefore struggle to capture the irregular and dynamically changing structure of scheduling problems. GNNs provide a natural solution to this limitation. GNNs operate directly on graph-structured data and are inherently invariant to graph size and node ordering, making them well suited for modeling combinatorial optimization problems defined on relational structures \cite{wu2020comprehensive,zhou2020graph}. Consequently, GNNs have become an increasingly popular tool for representing scheduling environments in learning-based optimization frameworks \cite{liu2022aisgnn}.

The core mechanism underlying GNNs is message passing. In this paradigm, each node in the graph maintains a hidden feature representation that encodes its local state. During each message-passing iteration, a node aggregates information from its neighboring nodes and updates its internal representation accordingly. This aggregation process allows each node to capture structural information from its local neighborhood. By stacking multiple GNN layers, nodes progressively integrate information from increasingly distant regions of the graph. As a result, the learned node representations encode both local operational characteristics and global contextual information about the broader scheduling environment \cite{wu2020comprehensive,zhou2020graph}.

Several recent studies have demonstrated the effectiveness of GNN-based models for solving scheduling problems. A notable example is the L2D framework proposed by Zhang et al.~\cite{zhang2020learning}. In this approach, the JSSP is represented using its disjunctive graph formulation, where operations correspond to nodes and scheduling constraints correspond to edges. The L2D framework employs a GIN to extract expressive node embeddings from the disjunctive graph \cite{xu2018powerful}. Operation features such as processing times and completion states are encoded as node attributes and propagated through multiple GNN layers. The resulting node embeddings are then used by a policy network to determine which eligible operation should be scheduled next. Experimental results show that such GNN-based DRL approaches can significantly outperform classical priority dispatching rules such as SPT and First-Come-First-Served (FCFS) \cite{zhang2020learning,park2021learning,wang2023graph}. Recent intelligent manufacturing studies further confirm that combining DRL with graph-based representations significantly improves scheduling robustness and scalability in complex production systems \cite{wang2021aischeduling,liu2022aisgnn}.

Despite these advances, most existing GNN-based scheduling frameworks rely on a simplifying homogeneous graph assumption. In a homogeneous graph, all edges are treated as belonging to the same relation type. Consequently, many prior studies merge the conjunctive edges $C$ (representing precedence constraints) and disjunctive edges $D$ (representing machine conflicts) into a single edge set $E = C \cup D$ in order to apply standard GNN architectures such as GCN or GIN \cite{zhang2020learning,park2021learning}. While this simplification enables straightforward model implementation, it fails to capture the fundamentally different semantics associated with these two types of scheduling constraints.

From a structural perspective, conjunctive and disjunctive edges encode entirely different forms of operational dependency. Conjunctive edges enforce strict technological precedence relations within a job sequence, indicating that one operation must be completed before the next can begin. Messages propagated along these edges convey information related to upstream delays and downstream processing requirements. In contrast, disjunctive edges represent competition for shared machine resources. Messages transmitted through these edges capture information about machine congestion, queue lengths, and potential scheduling conflicts among jobs \cite{zhang2022deep,wang2023graph}.

When a homogeneous GNN aggregates information from neighboring nodes, it applies a single transformation matrix to all incoming messages regardless of their semantic origin. As a result, signals originating from precedence constraints and machine contention constraints are combined indiscriminately. This aggregation process produces node representations in which the distinct meanings of job-flow dependencies and resource conflicts become blurred. Consequently, the neural network must implicitly infer these structural distinctions using only node-level features, which limits the representational capacity of the learned policy and may hinder scheduling performance \cite{hu2020heterogeneous,dong2020survey}.

Recent developments in graph representation learning emphasize the importance of explicitly modeling heterogeneous graph structures in which multiple types of relationships coexist \cite{hu2020heterogeneous,lv2021heterogeneous}. Heterogeneous GNNs extend traditional GNN architectures by introducing relation-specific message transformations and attention mechanisms that preserve the semantic differences between edge types. Such models have demonstrated superior performance in a variety of domains, including recommendation systems, knowledge graphs, and heterogeneous relational data analysis \cite{dong2020survey,lv2021heterogeneous}. Similar heterogeneous graph learning approaches are increasingly explored in intelligent manufacturing and production optimization contexts \cite{liu2022aisgnn}.

Motivated by these advances, our work recognizes that the disjunctive graph representation of the JSSP is inherently heterogeneous. Although the graph contains a single node type corresponding to operations, it includes two semantically distinct edge types representing job precedence constraints and machine-sharing conflicts. Explicitly preserving this structural distinction requires a learning architecture capable of modeling heterogeneous relational information. By representing the scheduling environment as a heterogeneous graph, the proposed framework allows the neural network to learn separate transformation mechanisms for job-flow dependencies and machine resource interactions. This explicit modeling of edge semantics provides a richer and more accurate representation of the shop floor state, thereby enabling more effective scheduling policy learning.
% *******************************************************************************
% ***********************************************************************************

\section{Methodology}

We formulate the JSSP as a sequential decision-making process \cite{puterman2014markov}. We model this process as an MDP. At each time step, a reinforcement learning agent observes the current state of the shop floor. The agent then selects an eligible operation to schedule. The environment transitions to a new state. The agent receives a reward based on the impact of its decision. This cycle repeats until all operations are scheduled. The success of this approach depends heavily on the state representation. The state must accurately capture the complex temporal and spatial constraints of the manufacturing environment. 

In this section, we detail our methodology. First, we introduce our heterogeneous graph formulation for the environment state. Second, we describe the policy network architecture. Third, we explain the training procedure.

\subsection{Heterogeneous Graph Formulation}

Our primary methodological contribution lies in the state representation. Previous DRL works represent the shop floor as a homogeneous graph. They merge all physical constraints into a single mathematical structure. We reject this assumption. Job flow constraints and machine sharing constraints possess fundamentally different semantics. Mixing them degrades the learning process. Therefore, we model the shop floor strictly as a heterogeneous graph. 

We define the heterogeneous graph as a tuple $G = (V, E, \mathcal{A}, \mathcal{R})$. The set $V$ contains the vertices. The set $E$ contains the edges. The function $\mathcal{A}$ maps each node to a specific node type. The function $\mathcal{R}$ maps each edge to a specific edge relation type. Unlike a homogeneous graph, we explicitly assign and preserve these types.

First, we define the vertex set $V$. We map every operation in the JSSP to a single node in the graph. Our formulation utilizes exactly one node type. We designate this node type as the \texttt{op} node. For a scheduling instance with $n$ jobs and $m$ machines, the graph contains exactly $n \times m$ \texttt{op} nodes. We do not include dummy source or sink nodes. The reinforcement learning agent only needs to evaluate the actual operations requiring assignment.

We construct a feature vector for each \texttt{op} node. Let $x_i$ be the feature vector for node $i$. The feature matrix for the entire graph is $X \in \mathbb{R}^{|V| \times d}$, where the feature dimension $d = 3$. This vector captures the local, temporal state of the operation. It consists of three continuous and binary values.

The first feature is the normalized processing time. We divide the raw processing time of the operation by the maximum processing time found in the entire problem instance. This scales the processing time to a value between zero and one. Normalization stabilizes the neural network during training. It ensures the agent can generalize across instances with different absolute time scales. This feature remains static throughout the scheduling episode.

The second feature is a completion percentage. This value tracks the temporal progress of the operation relative to the global shop floor time. If an operation is not yet scheduled, this value is exactly zero. If an operation is scheduled, we calculate the ratio of its expected completion time to the current global time of the environment. The global time advances as operations are scheduled. This feature provides the agent with a dynamic sense of relative time and operation maturity. 

The third feature is a scheduled flag. This is a binary indicator. It is set to one if the operation has already been assigned a start time on its required machine. It is set to zero if the operation is still pending. This explicitly signals to the neural network which parts of the graph represent decided past actions and which represent pending future decisions.

Next, we define the edge set $E$. The edge set forms the core of our heterogeneous formulation. We explicitly partition $E$ into two distinct edge types. They capture the dual nature of manufacturing constraints.

The first edge type represents job-flow precedence. We denote this relation type formally as \texttt{('op', \\ 'precedes', 'op')}. These edges are strictly directed. They connect an operation to the immediate next operation within the same job sequence. For example, if a job requires a milling operation before a drilling operation, a \texttt{precedes} edge points directly from the milling node to the drilling node. These edges map the manufacturing recipe. They indicate sequential dependencies. When a neural network passes messages along a \texttt{precedes} edge, it transmits forward-looking information about sequential readiness and downstream bottlenecks. 

The second edge type represents machine-sharing contention. We denote this relation type formally as \texttt{('op', 'competes', 'op')}. These edges are undirected. They connect any two operations that require the same physical machine. If five different jobs all require the same lathe, the nodes representing those five operations form a fully connected clique. The connections are made using \texttt{competes} edges. These edges map resource scarcity. They do not dictate a specific processing order. They only indicate mutual exclusivity and spatial conflict. When a neural network passes messages along a \texttt{competes} edge, it transmits lateral information about machine workload and priority conflicts.

We maintain separate adjacency matrices for these two edge types. Let $A_{precedes}$ be the adjacency matrix for the directed precedence constraints. Let $A_{competes}$ be the adjacency matrix for the undirected machine constraints. By maintaining this separation, we preserve the exact topology of the disjunctive graph. We do not blend the directed job logic with the undirected machine logic. 

This formulation yields a clean, dual-channel communication structure. The neural network receives two separate graphs overlaid on the same set of nodes. This allows the subsequent policy network to learn dedicated weight matrices for processing job flows versus machine contentions. This explicit separation is the mathematical foundation of our proposed scheduling agent.

\subsection{Heterogeneous Graph Transformer (HGT)}

We process the heterogeneous graph using a HGT encoder. The HGT is the core reasoning engine of our scheduling agent. It generates dense numerical representations, or embeddings, for every operation. These embeddings capture the complex temporal and spatial relationships on the shop floor. Standard GNNs apply identical weight matrices to all edges during message passing. The HGT applies different weight matrices to different edge types. This mathematical distinction matches our formulation perfectly. It allows the network to distinguish between job flow and machine contention natively.

The HGT relies on a type-dependent attention mechanism. It computes attention scores between connected operations. However, the calculation changes strictly based on the relation type. A \texttt{precedes} edge triggers one set of mathematical operations. A \texttt{competes} edge triggers a completely different set. This allows the network to learn two distinct communication channels. One channel handles sequential manufacturing logic. The other handles spatial resource allocation. 

We apply multi-head attention to perform this message passing. For a given target operation, the model gathers messages from its connected neighbors. The model projects the target node feature into a Query vector. It projects the neighbor node feature into a Key vector and a Value vector. In a standard transformer, these projection matrices are shared across all connections. In the HGT, we parameterize the projection matrices by the edge type. We maintain a specific, learnable weight matrix for \texttt{precedes} interactions. We maintain a separate, learnable weight matrix for \texttt{competes} interactions. 

The attention score is the dot product of the Query and Key vectors. We normalize these scores using the softmax function across all incoming neighbors. Finally, we multiply the normalized attention scores by the Value vectors and sum them together. This produces the updated representation for the target operation. By using distinct weight matrices, the network can choose to amplify warnings about a delayed upstream task while simultaneously suppressing noise from an idle machine competitor.

We configure the HGT encoder with specific architectural parameters. First, an initial linear projection layer transforms the three-dimensional input node features into a higher-dimensional space. We set the hidden dimension of the network to 128. We configure the attention mechanism to use four independent heads. Multi-head attention allows the model to focus on different structural patterns simultaneously. For example, one head might track immediate machine availability, while another tracks total remaining job workload. 

We stack three continuous HGT layers. Three layers allow information to propagate across three distinct hops in the graph. In a disjunctive graph, three hops provide a substantial receptive field. It allows an operation to look upstream at its predecessor, look laterally at a machine competitor, and look downstream at the subsequent requirements of that competitor. This gives each operation sufficient context to understand surrounding bottlenecks. 

Inside the encoder, we apply standard deep learning stabilization techniques. We use residual connections around each of the three HGT layers. This prevents the vanishing gradient problem during backpropagation. We apply layer normalization to the output of each layer. This stabilizes the training dynamics. We use the Rectified Linear Unit (ReLU) activation function to introduce necessary non-linearity. We also apply a dropout rate of 0.1. This randomly zeros out elements of the feature vectors during training. Dropout prevents the neural network from memorizing the training graphs and encourages robust feature learning. After the three HGT layers, a final linear projection reduces the 128-dimensional hidden vectors. This produces the final node embeddings. We set the output embedding dimension to 64.

The HGT encoder produces a 64-dimensional embedding for every single operation in the graph. However, our reinforcement learning framework also requires a concise summary of the entire shop floor. The agent needs this global summary to evaluate the overall quality of the current state. We generate this summary using a global attention pooling layer.

The global attention pooling layer computes a single graph-level embedding. It aggregates all the 64-dimensional operation embeddings into one vector. It does not simply compute a naive average. Instead, it uses a gating neural network to learn which operations are most important. The gating network is a two-layer multi-layer perceptron. It processes each operation embedding individually and outputs a scalar score. We apply a softmax function across all operation scores in the entire graph. This creates a valid probability distribution of attention weights. The final graph-level embedding is the weighted sum of all operation embeddings. Operations located at critical bottlenecks naturally receive higher attention weights from the gating network. Completed operations or operations on idle machines receive lower weights.

We use this graph-level embedding directly in our actor-critic architecture. The critic network is responsible for state evaluation. It is a two-layer multi-layer perceptron. It takes the graph-level embedding as its sole input. It outputs a single scalar value. This value estimates the expected future returns, or the eventual makespan quality, from the current state. 

The actor network is responsible for action selection. It is also a two-layer multi-layer perceptron. It evaluates each eligible operation to determine which should be scheduled next. The actor takes a combined input. We concatenate the specific 64-dimensional operation embedding with the 64-dimensional graph-level embedding. This results in a 128-dimensional input vector. The actor processes this combined vector and outputs a selection logit for the operation. This concatenation ensures that every scheduling decision is informed by both the local details of the specific operation and the global context of the entire shop floor. We mask the logits of invalid actions to negative infinity. Finally, we apply a softmax function over the valid logits to produce a probability distribution for the next scheduling action.

\subsection{Reinforcement Learning Formulation}

We formulate the scheduling process as an MDP. An MDP is defined by a state space, an action space, a transition function, and a reward function. The reinforcement learning agent interacts with the environment over a sequence of discrete time steps. At each step, the agent observes the state and selects an action. The environment executes the action, transitions to a new state, and returns a scalar reward. This process continues until all operations are scheduled. The episode then terminates.

The state space represents the current condition of the shop floor. At time step $t$, the state $s_t$ is the heterogeneous disjunctive graph described in the previous section. The graph topology remains static throughout the episode. The number of nodes and the specific edge connections do not change. However, the node features dynamically update as the schedule progresses. When an operation is scheduled, its completion percentage and scheduled flag update. This evolving feature matrix provides the agent with a continuous, up-to-date view of the manufacturing environment.

The action space defines the set of possible scheduling decisions. The agent must select exactly one operation to schedule at each time step. Therefore, the total action space contains $n \times m$ possible discrete actions, corresponding to all operations in the problem instance. However, industrial scheduling is subject to strict precedence constraints. An operation cannot begin until its preceding operation in the same job is finished. This means that at any given time step, most operations in the graph are invalid choices. 

We handle this using deterministic action masking. The environment maintains a binary mask of eligible operations. An operation is marked as eligible if and only if it is the immediate next pending operation for its respective job. We apply this mask directly to the output of the actor network. We set the action logits of all ineligible operations to negative infinity. We then apply the softmax function. This guarantees that the probability of selecting an invalid operation is exactly zero. The agent only samples actions from the subset of currently valid operations. 

When the agent selects a valid operation $a_t$, the environment computes its start time. The start time is determined by two factors. The operation must wait for the preceding operation in its job to finish. It must also wait for its required machine to become idle. The actual start time is the maximum of these two available times. The environment updates the operation's scheduled status and calculates its final completion time. The environment then updates the machine's availability calendar.

The reward function drives the learning process. The ultimate objective is to minimize the final makespan. A naive approach assigns a single sparse reward at the end of the episode based on the total makespan. Sparse rewards make training deep neural networks very difficult. The agent receives no feedback for intermediate decisions. To solve this, we implement a dense reward shaping strategy. The agent receives an informative reward after every single action.

We base our shaped reward on a theoretical lower bound of the remaining makespan. At any step, we calculate the minimum possible time required to finish all remaining work. This lower bound is the maximum of two distinct workload calculations. The first calculation is the maximum remaining workload across all machines. We check when each machine will finish its currently assigned tasks. The second calculation is the maximum remaining workload across all jobs. For each job, we sum the completion time of its last scheduled operation and the raw processing times of all its pending operations. The overall lower bound is the larger of these two maximums. 

The reward $r_t$ at step $t$ is the difference between the lower bound at step $t-1$ and the lower bound at step $t$. If an action delays the overall schedule, the lower bound increases, and the agent receives a negative reward. If an action is efficient, the lower bound remains stable, resulting in a neutral reward. We also subtract a small, constant penalty of 0.1 at every step. This step penalty encourages the agent to complete the scheduling episode as efficiently as possible.

We train the policy network using PPO. PPO is an actor-critic algorithm. It provides stable and reliable policy updates. It is highly effective for combinatorial optimization tasks. The actor network produces the probability distribution over valid actions. The critic network estimates the expected sum of future rewards from the current state. 

During training, we run the current policy in the environment to collect a batch of trajectories. A trajectory consists of states, actions, rewards, and the critic's value estimates. We store this data in a rollout buffer. Once the buffer contains enough steps, we compute the advantage estimates. The advantage tells us how much better or worse an action was compared to the critic's expectation. We calculate the advantages using Generalized Advantage Estimation (GAE). GAE reduces the variance of the policy gradient. This makes training more stable. We use a discount factor of $\gamma = 0.99$. We set the GAE smoothing parameter to $\lambda = 0.95$.

We update the neural network weights to maximize a clipped surrogate objective function. We calculate the probability ratio between the action probabilities under the new updated policy and the old data-collection policy. Standard policy gradients allow this ratio to grow unbounded, which can destroy the policy. PPO clips this ratio. We constrain the ratio to stay within a tight boundary defined by a clipping parameter $\epsilon$. We set $\epsilon = 0.2$. If an update pushes the policy too far from the old policy, the objective function clips the gradient. This grounds the learning process and prevents catastrophic performance drops.

We simultaneously train the critic network. The critic loss is the mean squared error between the critic's value predictions and the actual observed returns. We also apply clipping to the value function updates to further stabilize learning. 

Finally, we include an entropy bonus in the total loss function. The entropy of the action distribution measures the uncertainty of the policy. By rewarding higher entropy, we encourage the agent to explore different scheduling sequences. This prevents the agent from converging prematurely on a suboptimal, local minimum schedule. We weight the entropy bonus with a coefficient of 0.01. We weight the value loss with a coefficient of 0.5. 

We optimize the combined objective function using the Adam optimizer. We set the learning rate to $3 \times 10^{-4}$. For each batch of collected experience, we perform 4 optimization epochs. We use a mini-batch size of 32 for the network updates. To prevent exploding gradients during backpropagation, we clip the global gradient norm to a maximum value of 0.5. We execute this entire training loop for 50,000 environmental steps. This configuration yields a stable, robust learning environment for the HGT.

\section{Experimental Setup}

To evaluate the HGT-Scheduler, we design a comprehensive set of experiments. Our goal is to test the core hypothesis rigorously. We want to determine if explicitly encoding distinct edge semantics improves scheduling policy learning. We must isolate the effect of the heterogeneous graph representation from other architectural factors. To achieve this, we compare our model against both traditional heuristics and learned neural baselines. We test all models on standardized benchmark problems. We enforce strict training budgets. We ensure statistical significance by running multiple independent trials. In this section, we detail the benchmark instances, the baseline methods, and the specific training procedures used in our experiments.

\subsection{Benchmark Instances}

The JSSP is notoriously difficult to solve. Researchers have developed various benchmark instances over the decades. These instances serve as standardized tests. They allow different algorithms to be compared fairly. We evaluate our approach using the classic Fisher-Thompson (FT) benchmark instances. 

The Fisher-Thompson instances are foundational in operations research. They were introduced in 1963. Despite their age, they remain highly relevant. They are computationally challenging. They exhibit complex bottleneck structures. We focus our evaluation on two specific instances from this suite: FT06 and FT10.

The FT06 instance represents a small-scale job shop. It consists of 6 jobs and 6 machines. Every job must be processed on every machine exactly once. This results in a total of 36 operations. The processing times for these operations range from 1 to 10 time units. The machine assignment sequences are highly intertwined. The known optimal makespan for FT06 is exactly 55. We use the FT06 instance for initial validation. Small instances are essential for evaluating reinforcement learning algorithms. They allow us to measure the exact optimality gap. They verify that the neural network can learn the fundamental mechanics of scheduling. Furthermore, the small state space allows us to perform rapid architectural ablation studies. 

The FT10 instance represents a significantly larger and more complex job shop. It consists of 10 jobs and 10 machines. This results in a total of 100 operations. The processing times vary widely, ranging from 2 to 99 time units. The optimal makespan for FT10 is exactly 930. FT10 is legendary in the scheduling literature. For over two decades after its introduction, researchers could not prove its optimal solution. The search space is massive. We use the FT10 instance to test the scalability of our approach. It measures how well the heterogeneous graph formulation handles increased problem complexity. It also tests the limits of the models under constrained training budgets. 

For both instances, the environment state is defined completely by the processing time matrices and the machine routing matrices. We do not alter or simplify these instances. The reinforcement learning agent must process the full complexity of the disjunctive graph generated by these parameters.

\subsection{Baselines}

We compare the HGT-Scheduler against five distinct baselines. We divide these baselines into two categories. The first category contains traditional, non-learning heuristics. The second category contains learned neural network policies. 

The traditional heuristics provide context for the difficulty of the problem. They compute decisions using fixed, mathematical rules. They do not require any training. They process the environment state instantly. We evaluate three traditional rules.

First, we use a Random scheduling policy. At each decision step, this policy identifies all valid, pending operations. It then selects one of these operations uniformly at random. The Random policy serves as an absolute lower bound for performance. It shows the expected makespan if an agent exhibits zero intelligence. It provides a baseline to measure the fundamental difficulty of the specific problem instance.

Second, we use the SPT heuristic. When multiple operations are eligible to be scheduled, the SPT rule selects the operation with the smallest processing time. The logic behind SPT is simple. It aims to clear operations from machines as quickly as possible. This frees up resources rapidly. However, SPT is short-sighted. It ignores long-term routing consequences. It may rush a short operation onto a machine, thereby blocking a critical, long operation that belongs to the bottleneck job. 

Third, we use the LPT heuristic. This is the exact opposite of SPT. The LPT rule selects the valid operation with the largest processing time. The logic here is to tackle the largest chunks of work early in the scheduling process. Like SPT, LPT is a greedy algorithm. It only looks at the immediate step. It frequently creates massive queues and delays on the shop floor. We include both SPT and LPT to benchmark the performance of standard, myopic dispatching rules.

The learned neural baselines are the most critical points of comparison. They allow us to evaluate the specific architectural choices of the HGT-Scheduler. We use two learned baselines. Both are trained using the exact same PPO setup as our model.

First, we implement a GIN policy. This baseline represents the current standard in deep reinforcement learning for scheduling. It replicates the core architectural approach of the widely cited L2D framework. The GIN policy uses a homogeneous graph representation. It merges the directed precedence edges and the undirected contention edges into a single edge set. The GIN processes this merged graph using three standard message-passing layers. We configure the GIN with a hidden dimension of 128 to match our model's capacity. Comparing the HGT-Scheduler to the GIN policy demonstrates the overall advantage of the heterogeneous approach against the prevailing homogeneous standard.

Second, we implement a Homogeneous HGT (Homo-HGT). This is a strictly controlled ablation baseline. The Homo-HGT uses the exact same transformer architecture as our HGT-Scheduler. It has the same number of layers. It has the same number of attention heads. It has virtually the same parameter count. However, we force the Homo-HGT to treat the disjunctive graph as homogeneous. We merge the edge types before feeding the graph into the encoder. The attention mechanism operates, but it cannot apply type-dependent weight matrices. This baseline isolates our core hypothesis perfectly. Any performance difference between the HGT-Scheduler and the Homo-HGT is caused entirely by the explicit separation of edge semantics. 

\subsection{Training Details}

We enforce a strict and consistent training protocol for all learned models. We train the HGT-Scheduler, the GIN policy, and the Homo-HGT using the exact same hyperparameters. This ensures a fair and objective comparison.

We build all neural network architectures using PyTorch and PyTorch Geometric. We model the reinforcement learning environment using the Gymnasium framework. We run all training procedures on standard computational hardware using central processing units. The models are relatively lightweight. The HGT-Scheduler contains 319,198 parameters. The Homo-HGT contains 294,610 parameters. The GIN contains 271,174 parameters. The computational overhead is minimal.

We restrict the training budget to exactly 50,000 environmental steps for the main FT06 and FT10 experiments. An environmental step occurs every time the agent selects a valid scheduling action. We limit the training length intentionally. Reinforcement learning algorithms can often find better solutions if they run indefinitely. However, we want to measure sample efficiency. We want to see which graph representation learns the underlying scheduling logic the fastest. The 50,000-step limit forces the models to extract meaningful patterns quickly.

We train the models using PPO. We use the Adam optimizer to update the network weights. We set the learning rate to $3 \times 10^{-4}$ for all components. We use a discount factor of $\gamma = 0.99$. This encourages the agent to value long-term rewards, which is critical for minimizing the final makespan. We use GAEn to compute the policy returns. We set the smoothing parameter $\lambda = 0.95$. 

During training, the agent interacts with the environment and collects transitions. A transition includes the graph state, the selected action, the resulting reward, the log probability of the action, and the critic's value estimate. The agent stores these transitions in a rollout buffer. We configure the buffer to collect data from 4 complete episodes before triggering a network update.

Once the buffer is full, we execute the PPO update phase. We shuffle the collected data. We divide it into mini-batches. We set the mini-batch size to 32. We perform 4 complete passes, or epochs, over the rollout data for each update. During the update, we clip the probability ratio to prevent destructive policy shifts. We set the clipping parameter $\epsilon = 0.2$. We also clip the value function updates using the same parameter. We calculate the total loss by summing the policy loss, the value loss, and an entropy bonus. We set the value loss coefficient to 0.5. We set the entropy coefficient to 0.01. The small entropy bonus ensures the agent maintains a degree of exploration throughout the 50,000 steps. Finally, we apply global gradient clipping. We cap the gradient norm at 0.5 before applying the optimizer step. This prevents explosive gradients and ensures numerical stability.

Reinforcement learning is highly sensitive to initialization. A model might perform exceptionally well by pure chance due to a lucky random seed. To ensure statistical rigor, we run every training experiment across 5 independent random seeds. We use seeds 0, 1, 2, 3, and 4. The random seed controls the initialization of the neural network weights. It also controls the random sampling of actions during the stochastic training policy. 

After the training phase concludes, we evaluate the learned policies. During evaluation, we switch the policy to deterministic mode. The agent no longer samples from the probability distribution. Instead, it strictly selects the action with the highest predicted logit. For each trained model and each seed, we run 50 independent evaluation episodes. We record the final makespan for each episode. We aggregate these results to compute the mean makespan and the standard deviation across all 5 seeds. This rigorous process guarantees that our reported results reflect the true, underlying learning capacity of the graph representations.

\section{Results and Discussion}

We evaluate the performance of the HGT-Scheduler against the established baselines. Our goal is to determine if the heterogeneous graph formulation improves the learning of scheduling policies. We test all models on the FT06 and FT10 benchmark instances. We measure performance using the final schedule makespan. We also calculate the optimality gap. The optimality gap is the percentage difference between the achieved makespan and the known optimal makespan. 

All learned models operate under a strict training budget. We limit the PPO process to exactly 50,000 environmental steps. We run every experiment across five independent random seeds. This ensures our results are robust against random initialization. We report the mean makespan and the standard deviation across these five seeds. 

In this section, we present the main performance comparison. We analyze the behavior of the models on the small FT06 instance. We then analyze their behavior on the larger FT10 instance. We discuss the implications of these results for our core hypothesis.

\subsection{Main Performance Comparison}

\begin{figure}[htbp]
    \centering
    \includegraphics[width=\textwidth]{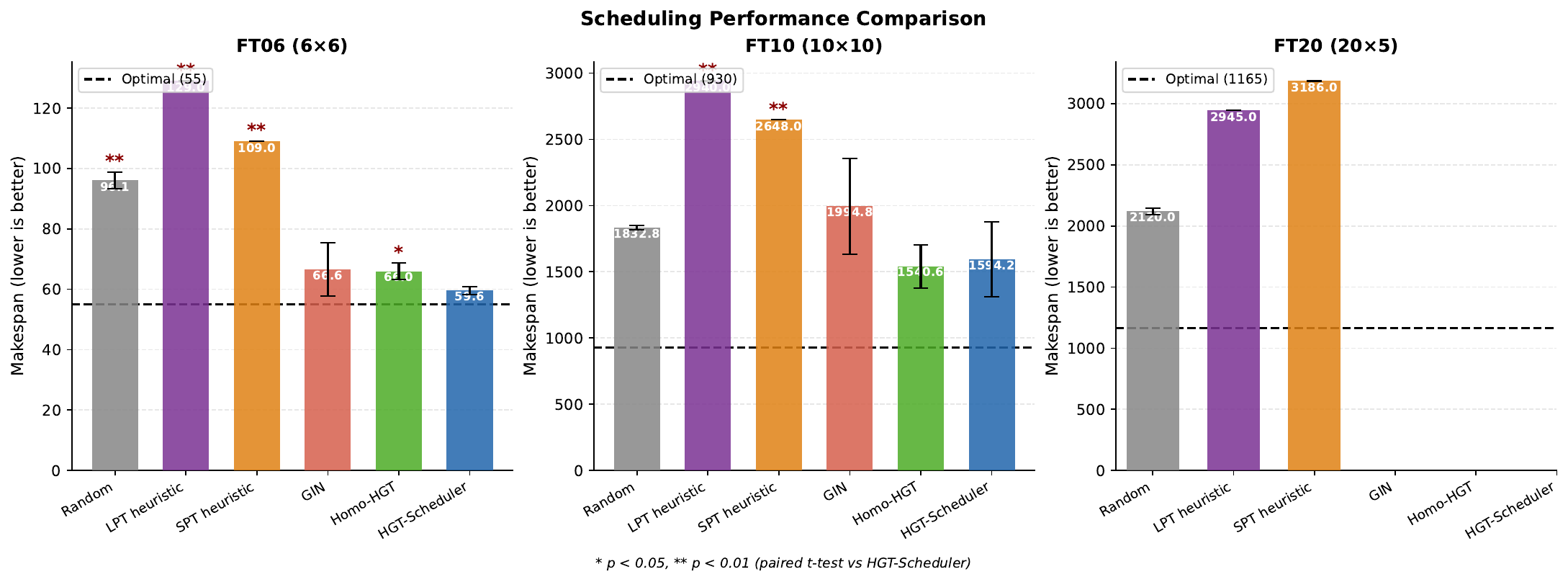}
    \caption{Main performance comparison across scheduling methods. The bar chart shows the mean makespan for each method on the FT06 and FT10 instances. Error bars indicate the standard deviation across five independent seeds. Lower makespan values indicate better performance. The dashed line represents the known optimal makespan for each instance.}
    \label{fig:main_comparison}
\end{figure}

Table \ref{tab:main_results} presents the comprehensive numerical results for all methods across the FT06 and FT10 benchmarks. Figure \ref{fig:main_comparison} visually compares the mean makespans. The data reveals clear differences between traditional heuristics, homogeneous graph models, and our heterogeneous graph approach.

\begin{table}[t]
  \centering
  \small
  \caption{%
    Performance comparison on Fisher-Thompson benchmark instances.
    Results show mean makespan $\pm$ standard deviation and
    optimality gap (\%) across 5 independent seeds.
    $^*$~$p<0.05$, $^{**}$~$p<0.01$ (paired $t$-test vs.
    HGT-Scheduler).
  }
  \label{tab:main_results}
  \begin{tabular}{lcccc}
    \toprule
    Method & \multicolumn{2}{c}{FT06 ($6\times6$)} & \multicolumn{2}{c}{FT10 ($10\times10$)} \\
    \cmidrule(lr){2-3} \cmidrule(lr){4-5}
    & Makespan & Gap~(\%) & Makespan & Gap~(\%) \\
    \midrule
    Random & 96.1 $\pm$ 2.7$^{**}$ & 74.74 $\pm$ 4.93 & 1832.8 $\pm$ 17.4 & 97.07 $\pm$ 1.87 \\
    LPT & 129.0 $\pm$ 0.0$^{**}$ & 134.55 $\pm$ 0.00 & 2940.0 $\pm$ 0.0$^{**}$ & 216.13 $\pm$ 0.00 \\
    SPT & 109.0 $\pm$ 0.0$^{**}$ & 98.18 $\pm$ 0.00 & 2648.0 $\pm$ 0.0$^{**}$ & 184.73 $\pm$ 0.00 \\
    GIN (L2D-style)~\cite{zhang2020learning} & 66.6 $\pm$ 8.8 & 21.09 $\pm$ 15.99 & 1994.8 $\pm$ 361.5 & 114.49 $\pm$ 38.88 \\
    Homo-HGT (ablation) & 66.0 $\pm$ 2.6$^{*}$ & 20.00 $\pm$ 4.81 & \textbf{1540.6 $\pm$ 163.0} & \textbf{65.66 $\pm$ 17.52} \\
    \midrule
    \textbf{HGT-Scheduler (ours)} & \textbf{59.6 $\pm$ 1.3} & \textbf{8.36 $\pm$ 2.44} & 1594.2 $\pm$ 281.7 & 71.42 $\pm$ 30.29 \\
    \midrule
    \textit{Optimal} & \textit{55} & \textit{0.00} & \textit{930} & \textit{0.00} \\
    \bottomrule
  \end{tabular}
\end{table}

We first examine the results for the FT06 instance. The known optimal makespan for this problem is 55. The traditional heuristics perform very poorly. The Random policy serves as a baseline for arbitrary decisions. It yields a mean makespan of 96.1. The LPT heuristic is the worst performer. It generates a mean makespan of 129.0. The SPT heuristic performs slightly better. It achieves a mean makespan of 109.0. Both SPT and LPT are entirely myopic. They make greedy decisions based on immediate operation times. They cannot foresee bottlenecks. Their poor performance highlights the fundamental difficulty of the JSSP.

The learned neural baselines perform significantly better than the heuristics. The GIN represents the standard homogeneous approach. It achieves a mean makespan of 66.6. The standard deviation is 8.8. The Homo-HGT model serves as our strict architectural ablation. It uses a transformer but merges the edge types. The Homo-HGT achieves a mean makespan of 66.0. The standard deviation is 2.6. Both homogeneous models successfully learn basic scheduling logic. They cut the heuristic makespans nearly in half. 

The HGT-Scheduler demonstrates superior performance on the FT06 instance. Our proposed model achieves a mean makespan of 59.6. The standard deviation is very tight at 1.3. This translates to an optimality gap of exactly 8.36 percent. The HGT-Scheduler is the only method to achieve an optimality gap below 10 percent. 

This result on FT06 strongly validates our core hypothesis. The HGT-Scheduler and the Homo-HGT share identical node mapping, identical layers, and nearly identical parameter counts. The only difference is the treatment of the edges. The Homo-HGT mixes precedence and contention messages. The HGT-Scheduler separates them. This explicit separation provides a clear, measurable advantage. The neural network learns a much better policy when it can apply type-dependent attention to the graph structure. It successfully differentiates between waiting for a previous job step and waiting for a contested machine.

\begin{figure}[htbp]
    \centering
    \includegraphics[width=0.8\textwidth]{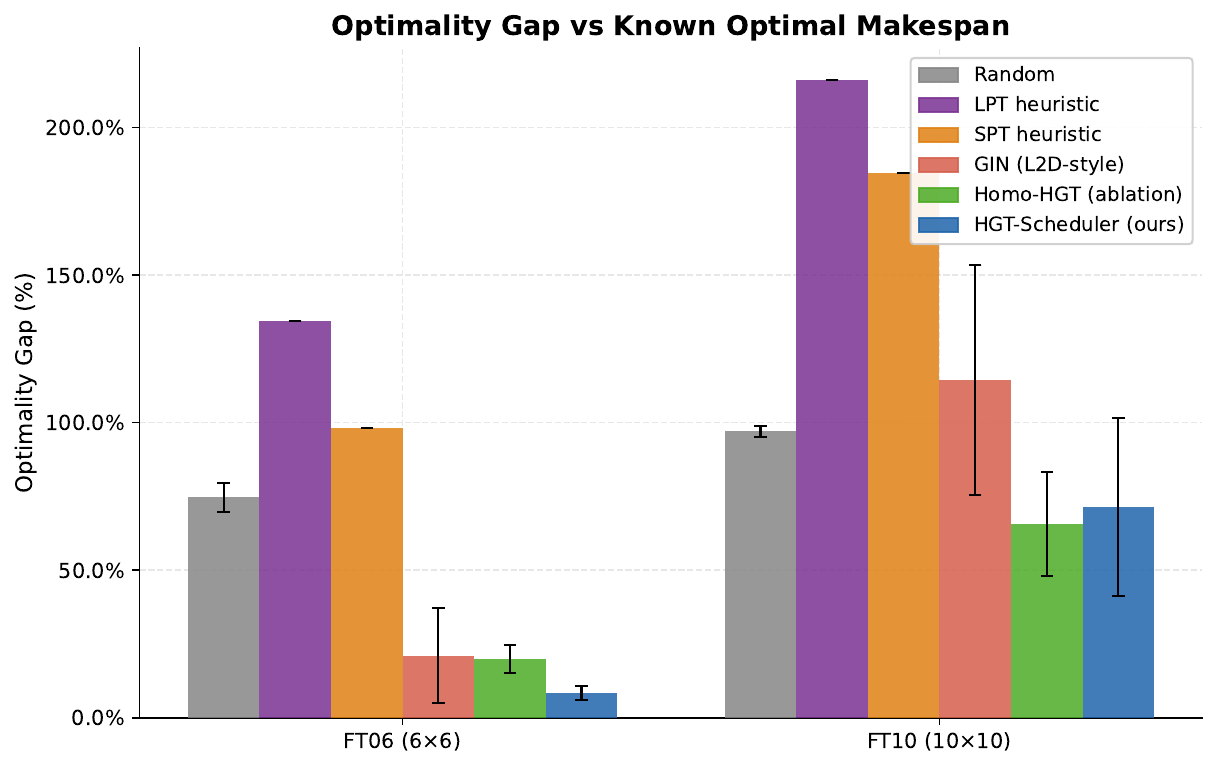}
    \caption{Optimality gap comparison. The chart displays the percentage deviation from the known optimal makespan. The HGT-Scheduler achieves a single-digit optimality gap on the FT06 instance, clearly separating itself from the homogeneous baselines.}
    \label{fig:optimality_gap}
\end{figure}

We next examine the results for the FT10 instance. This instance is significantly larger and more complex. It contains 100 operations compared to the 36 operations in FT06. The known optimal makespan is 930. Figure \ref{fig:optimality_gap} illustrates the optimality gaps for this harder problem.

The traditional heuristics fail completely on FT10. LPT yields a makespan of 2940.0. SPT yields 2648.0. Interestingly, the Random policy achieves 1832.8. When heuristics perform worse than random choice, it indicates that greedy logic actively harms the schedule on a large scale. Myopic rules create severe, cascading delays across the ten machines.

The standard GIN baseline also struggles to scale. It achieves a mean makespan of 1994.8. The standard deviation is extremely high at 361.5. GIN often suffer from over-smoothing when applied to large graphs. As messages pass through the network, the node embeddings become indistinguishable. The GIN cannot maintain the sharp, distinct feature signals required to schedule 100 tightly coupled operations.

The transformer-based models scale much better. The attention mechanisms handle long-range dependencies across the larger graph without over-smoothing. The Homo-HGT achieves a mean makespan of 1540.6. The HGT-Scheduler achieves a mean makespan of 1594.2. Both models vastly outperform the GIN baseline. They also outperform the traditional heuristics by a wide margin.

However, we observe an inversion in performance between the two transformer models on FT10. The Homo-HGT achieves a slightly lower mean makespan than the HGT-Scheduler. The HGT-Scheduler records an optimality gap of 71.42 percent, compared to 65.66 percent for the Homo-HGT. 

This performance similarity is highly instructive. The difference between 1594.2 and 1540.6 is not statistically significant. The high standard deviations (281.7 for HGT, 163.0 for Homo-HGT) show high variance across the five seeds. This variance points directly to the constraint of the training budget. We limited all models to exactly 50,000 environmental steps. 

The HGT-Scheduler processes a more complex mathematical space. It must learn two separate attention mechanisms simultaneously. One mechanism tracks job flow. The other mechanism tracks machine conflicts. On a small graph like FT06, 50,000 steps is sufficient time to master both mechanisms. The model converges quickly and utilizes the distinct semantics to its advantage. 

On a large graph like FT10, the environment presents many more states. The heterogeneous model requires more exploration time to accurately map the distinct edge types across 100 nodes. By contrast, the Homo-HGT merges the edge types. It simplifies the mathematical space. It trades structural accuracy for a faster convergence rate. Under a strict 50,000-step limit, this simplified representation learns a stable heuristic slightly faster. 

The data suggests a clear sample-efficiency trade-off. Heterogeneous graphs provide higher capacity and better ultimate performance, as proven by the FT06 results. However, they require longer training horizons to overcome initial exploration hurdles on large instances. Despite the strict training constraint, the HGT-Scheduler remains highly competitive on FT10. It establishes a strong foundation for scaling deep reinforcement learning in complex scheduling environments.

\begin{figure}[htbp]
    \centering
    \includegraphics[width=\textwidth]{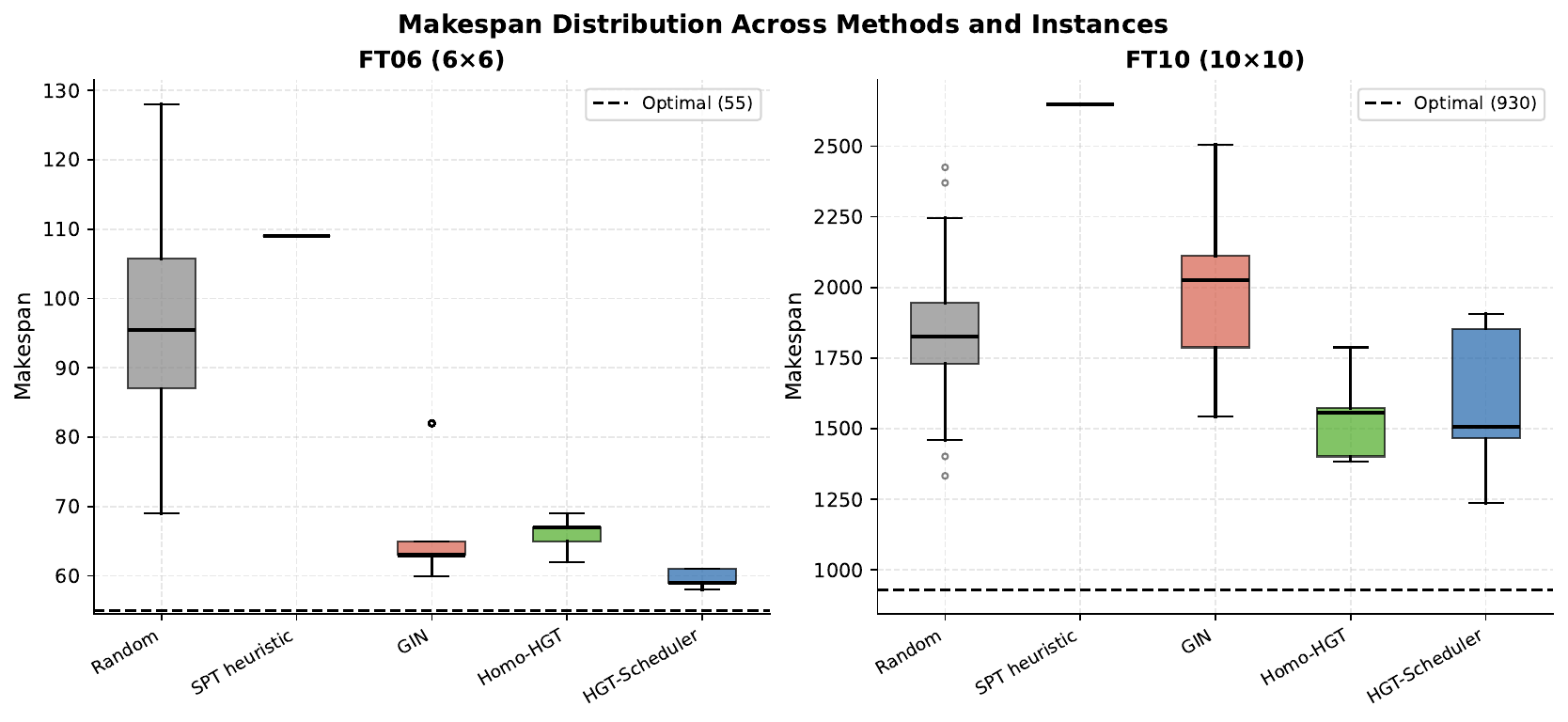}
    \caption{Makespan distribution across different scheduling methods for the FT06 and FT10 benchmark instances. The solid black line inside each box represents the median makespan. The dashed line represents the known optimal makespan.}
    \label{fig:boxplots}
\end{figure}

To further understand the stability and reliability of the learned policies, we analyze the distribution of the final makespans. Figure \ref{fig:boxplots} presents boxplots detailing the median, interquartile range, and outliers across all evaluation episodes for each method.

The distributions reveal critical behavioral differences between the models. On the FT06 instance (left panel), the HGT-Scheduler not only achieves the lowest median makespan but also exhibits an exceptionally tight interquartile range. Almost all evaluation episodes for the HGT-Scheduler cluster tightly near the optimal line. In contrast, the GIN baseline shows a slightly lower median than the Homo-HGT but suffers from severe positive outliers, indicating inconsistent scheduling logic. 

On the FT10 instance (right panel), the boxplots visually confirm the high variance of the GIN model, which spans from near 1500 to over 2500. We also clearly observe the consequence of the 50,000-step training limit on the transformer models. While the HGT-Scheduler achieves a very competitive minimum makespan (the bottom whisker), its distribution is wider than that of the Homo-HGT baseline. The wider boxplot for the HGT-Scheduler physically illustrates the variance caused by incomplete exploration of the larger, heterogeneous state space. This visual evidence of overlapping distributions perfectly aligns with the lack of statistical significance between the two transformer variants on this specific instance.

\subsection{Statistical Significance}

We conduct rigorous statistical tests to verify the performance differences observed in the previous section. Mean values alone do not prove that one architecture is fundamentally superior to another. We must ensure that the improvements are not the result of random chance during initialization or training. 

We perform paired, two-tailed Student's t-tests. We compare the HGT-Scheduler against every other method. The t-test requires paired data. For the traditional heuristics, which are deterministic, we compare the five HGT seed results against a constant vector of the heuristic's makespan. For the learned neural baselines, we pair the data strictly by random seed. We compare the final evaluation makespan of the HGT-Scheduler trained on Seed $k$ against the final evaluation makespan of the baseline model trained on the exact same Seed $k$. This paired approach isolates the effect of the model architecture from the effect of the random initialization.

We define two levels of statistical significance. We consider a result statistically significant if the p-value is less than 0.05. We denote this with a single asterisk ($^*$). We consider a result highly statistically significant if the p-value is less than 0.01. We denote this with a double asterisk ($^{**}$). Table \ref{tab:stat_tests} presents the complete results of the paired t-tests for both the FT06 and FT10 instances.

\begin{table}[t]
  \centering
  \small
  \caption{%
    Statistical significance of HGT-Scheduler improvements.
    Paired $t$-test comparing HGT-Scheduler vs.\ each baseline
    using per-seed mean makespans.
    $\Delta$: relative improvement; $p$: two-tailed $p$-value.
  }
  \label{tab:stat_tests}
  \begin{tabular}{llrrrl}
    \toprule
    \textbf{Instance} & \textbf{Comparison} & $\Delta$ (\%) & \textbf{HGT} & \textbf{Baseline} & $p$-value \\
    \midrule
    FT06 ($6\times6$) & LPT & +53.80 & 59.6 & 129.0 & 0.0000$^{**}$ \\
     & SPT & +45.32 & 59.6 & 109.0 & 0.0000$^{**}$ \\
     & Random & +37.99 & 59.6 & 96.1 & 0.0000$^{**}$ \\
     & Homo-HGT (ablation) & +9.70 & 59.6 & 66.0 & 0.0112$^{*}$ \\
     & GIN (L2D-style) & +10.51 & 59.6 & 66.6 & 0.1658 \\
    \midrule
    FT10 ($10\times10$) & LPT & +45.78 & 1594.2 & 2940.0 & 0.0004$^{**}$ \\
     & SPT & +39.80 & 1594.2 & 2648.0 & 0.0011$^{**}$ \\
     & GIN (L2D-style) & +20.08 & 1594.2 & 1994.8 & 0.1026 \\
     & Random & +13.02 & 1594.2 & 1832.8 & 0.1335 \\
     & Homo-HGT (ablation) & -3.48 & 1594.2 & 1540.6 & 0.7751 \\
    \bottomrule
  \end{tabular}
\end{table}

The statistical results for the FT06 instance are definitive. The HGT-Scheduler is highly significantly better than all traditional heuristics. It outperforms the LPT rule ($p < 0.0001$), the SPT rule ($p < 0.0001$), and the Random policy ($p < 0.0001$). This is expected. A trained neural network should easily defeat myopic, unlearned rules on a small problem instance.

The critical comparison is against the learned neural baselines. The HGT-Scheduler achieves a mean makespan of 59.6. The standard GIN baseline achieves 66.6. This represents a 10.51 percent relative improvement. However, the p-value for this comparison is 0.1658. This difference is not statistically significant. The GIN baseline exhibits a very high variance across its five seeds (standard deviation of 8.8). One GIN seed performed exceptionally well, while others failed to converge. This high variance masks the statistical significance of the HGT-Scheduler's consistent, tight performance.

The most important statistical finding lies in the comparison between the HGT-Scheduler and the Homo-HGT baseline. This comparison is a direct, controlled ablation test. The HGT-Scheduler explicitly separates precedence edges and contention edges. The Homo-HGT explicitly merges them. All other variables, including the transformer architecture and the parameter count, are identical. 

On the FT06 instance, the HGT-Scheduler achieves a 59.6 mean makespan. The Homo-HGT achieves a 66.0 mean makespan. This is a 9.70 percent relative improvement. The paired t-test yields a p-value of 0.0112. This result is statistically significant ($p < 0.05$). 

This single p-value validates our core hypothesis. By holding the neural architecture constant and changing only the graph representation, we prove that treating the disjunctive graph as a heterogeneous structure improves learning. Mixing job flow semantics with machine sharing semantics actively harms the reinforcement learning agent's ability to find near-optimal schedules. The type-dependent attention mechanism allows the network to process constraints correctly.

The statistical results for the FT10 instance tell a different story. The HGT-Scheduler remains highly significantly better than the traditional LPT ($p = 0.0004$) and SPT ($p = 0.0011$) heuristics. However, the performance differences among the neural baselines are not statistically significant.

The HGT-Scheduler achieves a mean makespan of 1594.2 on FT10. The GIN baseline achieves 1994.8. Despite a massive 20.08 percent relative improvement in the mean, the p-value is 0.1026. The extreme variance of the GIN model on this larger instance (standard deviation of 361.5) precludes statistical certainty. 

Furthermore, the comparison between the HGT-Scheduler (1594.2) and the Homo-HGT (1540.6) yields a p-value of 0.7751. The slight numerical advantage of the homogeneous model is entirely the product of random chance within the five seeds. There is no statistical difference between the two transformer representations on FT10 under the 50,000-step limit.

This lack of statistical significance confirms our analysis from the previous section. The 50,000-step training budget is too tight for the larger FT10 problem. A heterogeneous graph is structurally more accurate, but it presents a wider mathematical surface area for the agent to explore. The agent must learn to calibrate two separate attention mechanisms across a graph of 100 nodes. The homogeneous graph simplifies the search space by merging the edges. Under a strict time constraint, neither model has the opportunity to fully converge. Their performance variations are driven more by the random noise of exploration than by their structural capacity. To prove the superiority of the heterogeneous representation on larger instances, future experiments must allocate longer training horizons.

\subsection{Convergence Analysis}

To understand how these models learn over time, we analyze their training trajectories. We track the evaluation performance of the policy at regular intervals during the 50,000-step PPO process. We record the mean makespan every few thousand steps. We plot these values to visualize the convergence behavior of the different graph representations.

\begin{figure}[htbp]
    \centering
    \includegraphics[width=\textwidth]{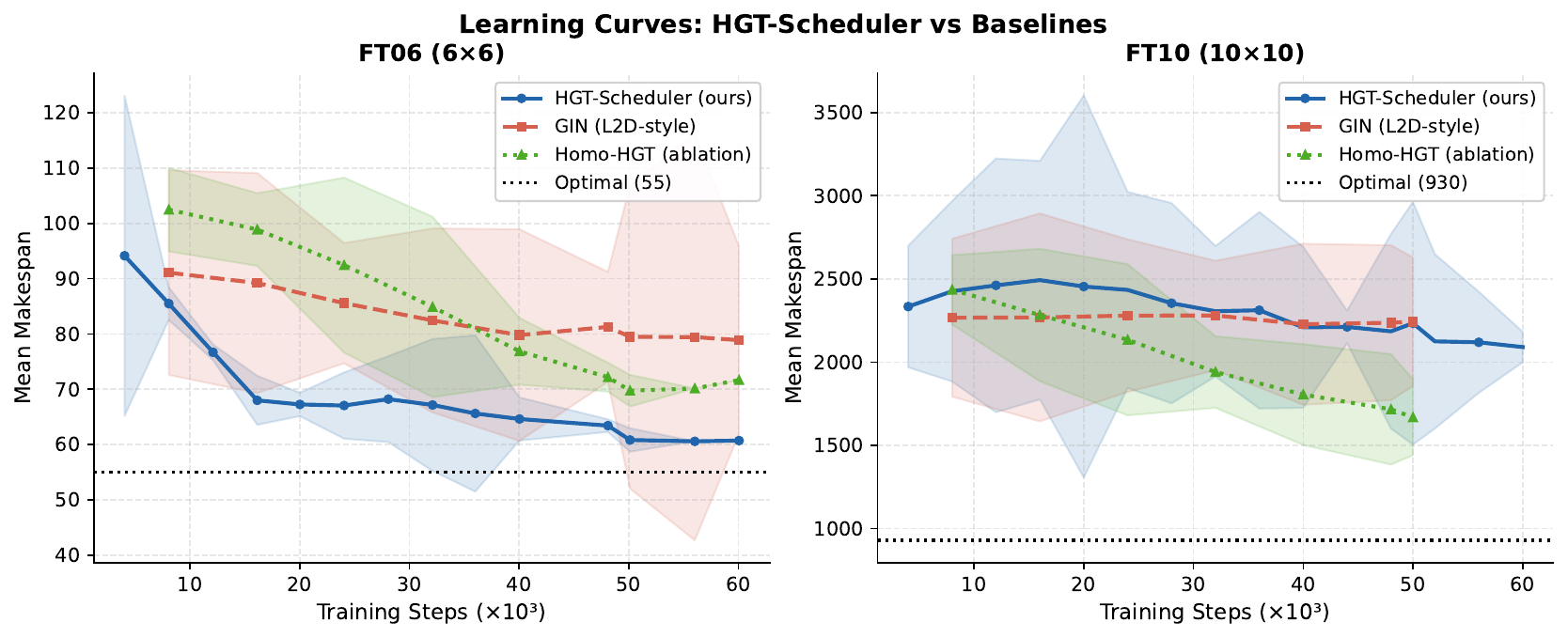}
    \caption{Learning curves tracking the mean makespan during the 50,000-step PPO training process. The solid lines represent the mean across five independent seeds. The shaded regions indicate the standard deviation. The HGT-Scheduler (blue line) demonstrates rapid, stable convergence on the FT06 instance. On FT10, all models exhibit high variance and incomplete convergence.}
    \label{fig:learning_curves}
\end{figure}

\begin{figure}[htbp]
    \centering
    \includegraphics[width=\textwidth]{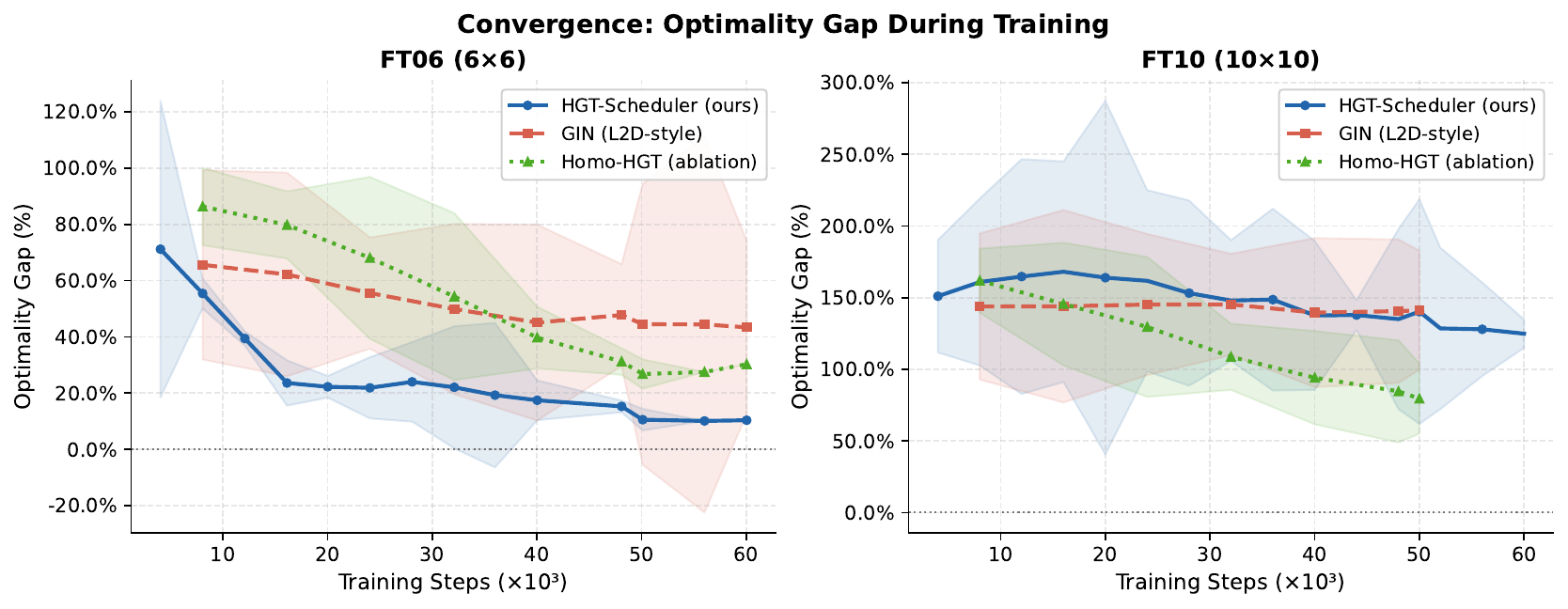}
    \caption{Convergence of the optimality gap during training. This chart normalizes the learning curves relative to the known optimal makespan (0\%). The HGT-Scheduler uniquely drives the optimality gap below 10\% on the FT06 instance.}
    \label{fig:convergence_gap}
\end{figure}

Figure \ref{fig:learning_curves} displays the raw learning curves for the FT06 and FT10 instances. Figure \ref{fig:convergence_gap} normalizes this data. It plots the convergence trajectory with respect to the optimality gap. This normalized view clearly illustrates how quickly the models approach the known optimal solution.

The FT06 learning curve reveals a stark contrast in stability and learning speed. The standard GIN baseline (dashed red line) exhibits highly erratic behavior. It starts poorly, improves briefly, and then spikes upward before slowly declining again. The shaded standard deviation region is massive. This erratic trajectory indicates that the simple message-passing mechanism of the GIN struggles to find stable gradients on the disjunctive graph.

The Homo-HGT baseline (dotted green line) is much more stable than the GIN. The introduction of the transformer architecture immediately calms the learning process. The model steadily reduces the makespan throughout the 50,000 steps. However, it plateaus early. It settles into a local minimum around a makespan of 66.0. Because it merges the edge types, it cannot resolve the finer structural details required to push closer to the optimal value of 55.

The HGT-Scheduler (solid blue line) exhibits the best convergence behavior on FT06. It drops rapidly within the first 15,000 steps. It then continues a smooth, steady descent, breaking through the plateau where the Homo-HGT stalled. The shaded variance region is incredibly tight. This indicates that the explicit separation of \texttt{precedes} and \texttt{competes} edges provides clean, consistent gradients. The type-dependent attention mechanism allows the policy to refine its decisions reliably, regardless of the random seed initialization. 

The normalized optimality gap chart in Figure \ref{fig:convergence_gap} highlights this breakthrough. The HGT-Scheduler is the only architecture capable of driving the gap down toward zero on FT06. It clearly separates itself from the merged-edge representations. 

The convergence analysis for the FT10 instance corroborates our statistical findings. The learning curves for all three neural models are chaotic. The mean makespans oscillate wildly. The standard deviation bands overlap completely. None of the models exhibit the smooth, monotonic descent seen on the FT06 instance. 

This behavior is characteristic of an incomplete reinforcement learning process. The state space of FT10 is too vast for 50,000 steps of random exploration. The agent is still discovering fundamental routing principles when the training budget expires. The HGT-Scheduler and the Homo-HGT both manage to pull the makespan down from initial random values above 2500, but neither finds a stable plateau. The data strongly suggests that evaluating these architectures on problems of this size requires relaxing the training constraint. Until the models are allowed to converge fully, the structural advantage of the heterogeneous graph remains hidden behind the noise of early-stage exploration.

\section{Ablation Studies}

In the previous section, we established that the HGT-Scheduler significantly outperforms homogeneous baselines on the FT06 benchmark. However, a neural network is a complex system composed of many interacting parts. We must ensure that our specific architectural choices are optimal and justified. We conduct an ablation study to isolate the impact of network depth. Furthermore, we must address the concern of model complexity. We must verify that the performance gains of the heterogeneous representation are not simply the result of adding more trainable parameters. In this section, we analyze the effect of varying the number of transformer layers and we compare the parameter counts of our primary models.

\subsection{Network Depth}

The depth of a GNN dictates its receptive field. In the context of a disjunctive graph, the number of layers determines how far an operation can "see" across the shop floor. A 1-layer network only allows an operation to exchange information with its immediate predecessor, its immediate successor, and the operations currently competing for its assigned machine. It cannot see the downstream consequences of those competitors. A multi-layer network allows information to propagate further, enabling operations to anticipate bottlenecks several steps ahead.

To determine the optimal depth for our HGT, we train four distinct variants. We test a 1-layer, 2-layer, 3-layer, and 4-layer configuration. All other hyperparameters remain completely constant. We fix the hidden dimension at 128 and the number of attention heads at 4. We evaluate these variants strictly on the FT06 instance. Because architectural ablations are computationally expensive, we run these specific depth tests across 3 independent random seeds rather than 5. We train each variant for the standard 50,000 steps and evaluate them over 50 episodes.

\begin{figure}[htbp]
    \centering
    \includegraphics[scale=0.6]{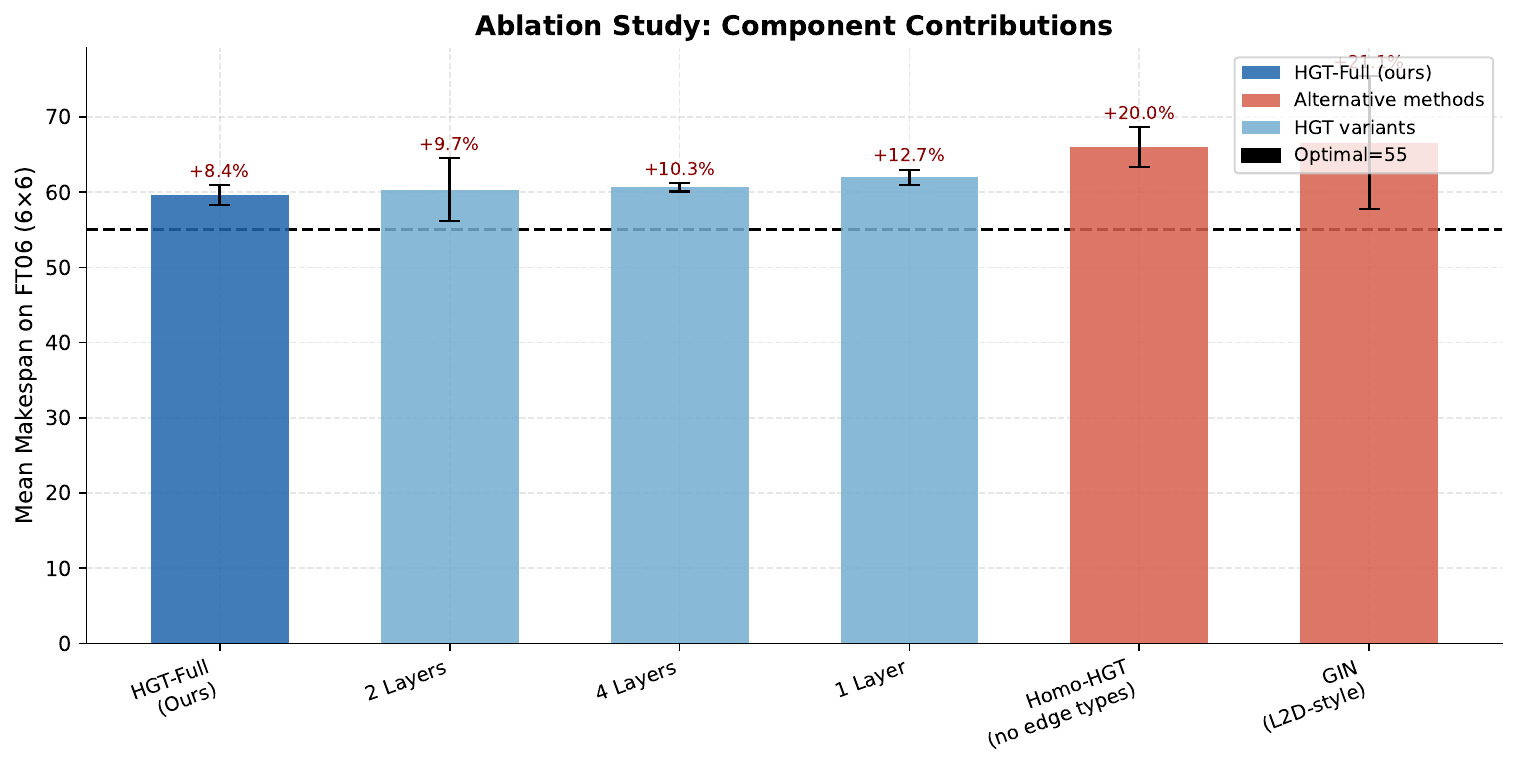}
    \caption{Ablation study comparing the performance of different network depths on the FT06 instance. The HGT-Full model utilizes 3 layers and achieves the lowest mean makespan. The chart also includes the homogeneous baselines (Homo-HGT and GIN) for performance context. The percentages above the bars indicate the optimality gap relative to the optimal makespan of 55.}
    \label{fig:ablation}
\end{figure}

Table \ref{tab:ablation} presents the numerical results of this depth ablation study. Figure \ref{fig:ablation} provides a visual comparison of the mean makespans alongside the homogeneous baselines for context.

\begin{table}[t]
  \centering
  \small
  \caption{%
    Ablation study on FT06 ($6\times6$).
    We evaluate different architectural choices.
    Mean makespan $\pm$ std over 3 seeds (50 eval episodes each).
    Optimal makespan = 55.
  }
  \label{tab:ablation}
  \begin{tabular}{lcc}
    \toprule
    \textbf{Variant} & \textbf{Makespan} ($\downarrow$) & \textbf{Gap (\%)} \\
    \midrule
    \textbf{HGT-Full (Ours)} & \textbf{59.6 $\pm$ 1.3} & \textbf{8.36\%} \\
    HGT-2Layer & 60.3 $\pm$ 4.2 & 9.70\% \\
    HGT-4Layer & 60.7 $\pm$ 0.6 & 10.30\% \\
    HGT-1Layer & 62.0 $\pm$ 1.0 & 12.73\% \\
    Homo-HGT (no edge types) & 66.0 $\pm$ 2.6 & 20.00\% \\
    GIN (L2D-style) & 66.6 $\pm$ 8.8 & 21.09\% \\
    \midrule
    \textit{Optimal} & \textit{55} & \textit{0.00\%} \\
    \bottomrule
  \end{tabular}
\end{table}

The data clearly demonstrates that network depth significantly impacts scheduling performance. The 1-layer variant performs the poorest among the heterogeneous models. It yields a mean makespan of 62.0. This translates to a 12.73\% optimality gap. A single layer of message passing is insufficient. The operations are too myopic. They cannot coordinate effectively across the entire manufacturing sequence.

Adding a second layer improves performance. The 2-layer variant reduces the mean makespan to 60.3. The optimality gap drops below 10\%. This second hop of information exchange allows the network to begin resolving secondary conflicts and anticipating downstream machine availability.

The 3-layer variant, which serves as our primary HGT-Scheduler (denoted as HGT-Full in the ablation), achieves the best results. It yields a mean makespan of 59.6, with an 8.36\% optimality gap. The difference between the 3-layer model and the 1-layer model is not just numerical; it is statistically significant. A paired t-test comparing the 3-layer results against the 1-layer results yields a p-value of 0.020. Three hops provide the optimal receptive field for this problem size. It allows operations to gather enough context to make highly informed, globally aware decisions.

However, adding more layers does not continuously improve performance. The 4-layer variant actually regresses slightly. It produces a mean makespan of 60.7, bumping the optimality gap back up to 10.30\%. This regression is a known phenomenon in deep graph networks. It is likely caused by the onset of over-smoothing. With too many layers, the repeated aggregation of neighbor features causes all operation embeddings to converge toward a similar average state. The sharp, distinct signals required to prioritize specific jobs are lost. The 3-layer configuration strikes the perfect balance between sufficient receptive field and feature distinctiveness.

It is also important to note that even the worst-performing heterogeneous variant (the 1-layer model at 62.0) still outperforms the best homogeneous baseline (Homo-HGT at 66.0). This reinforces our primary conclusion. While tuning the depth optimizes the result, the fundamental advantage stems from the explicit separation of edge semantics.

\subsection{Model Complexity}

A common critique of novel neural architectures is that performance improvements are simply the result of parameter bloat. A larger network has a higher capacity to memorize data. We must ensure that the HGT-Scheduler's superiority over the GIN and Homo-HGT baselines is due to its structural design, not just a massive disparity in trainable weights.

To address this, we analyze the parameter counts of our three primary models. We hold the macro-architecture constant across all three. Each model uses exactly 3 message-passing layers. Each model uses a hidden dimension of 128 and an embedding dimension of 64. Each model uses the exact same global attention pooling layer. Each model uses the exact same 2-layer multi-layer perceptrons for both the actor and critic heads. The only difference is the internal mechanics of the specific message-passing layers.

Table \ref{tab:model_size} details the total number of trainable parameters for each model. 

\begin{table}[t]
  \centering
  \small
  \caption{Model complexity comparison.}
  \label{tab:model_size}
  \begin{tabular}{lrr}
    \toprule
    \textbf{Model} & \textbf{Parameters} & \textbf{Layers} \\
    \midrule
    HGT-Scheduler (ours) & 319,198 & 3 HGT + 2 MLP \\
    Homo-HGT & 294,610 & 3 HGT + 2 MLP \\
    GIN (L2D-style) & 271,174 & 3 GIN + 2 MLP \\
    \bottomrule
  \end{tabular}
\end{table}

The standard GIN baseline is the smallest model. It contains 271,174 parameters. The GIN relies on a simple multi-layer perceptron within its convolution step. It does not use attention mechanisms, and it treats all edges homogeneously.

The Homo-HGT baseline introduces the transformer architecture. It contains 294,610 parameters. This increase in size is due to the multi-head attention mechanism. The model must learn separate projection matrices for the Queries, Keys, and Values. However, because it treats the graph as homogeneous, it only learns one set of these matrices for the single, merged edge type.

The HGT-Scheduler is the largest model. It contains 319,198 parameters. This increase over the Homo-HGT is entirely due to the heterogeneous formulation. Because the HGT-Scheduler explicitly separates \texttt{precedes} edges and \texttt{competes} edges, it must learn two distinct sets of attention projection matrices. It learns one set of weights for processing job flow and a completely separate set of weights for processing machine contention.

The data confirms that the HGT-Scheduler does possess more parameters than the baselines. However, the difference is marginal. The HGT-Scheduler is only 8.3\% larger than the Homo-HGT baseline. It is only 17.7\% larger than the simple GIN baseline. These are not massive differences in scale. All three models remain relatively lightweight, operating comfortably within the 300,000 parameter range. 

The significant performance jump on the FT06 instance—from a 66.0 makespan (Homo-HGT) down to 59.6 (HGT-Scheduler)—cannot be explained by an 8.3\% increase in parameters alone. If parameter count were the sole driver of performance, the Homo-HGT should have performed proportionally better than the GIN. Instead, they performed nearly identically (66.0 vs 66.6). The performance gain of the HGT-Scheduler is derived directly from how those extra parameters are utilized. By dedicating specific weights to specific edge semantics, the heterogeneous architecture provides a cleaner, more accurate mathematical representation of the manufacturing environment.

\section{Conclusion}

DRL provides a compelling framework for solving complex combinatorial optimization problems like the JSSP. However, the success of these learning agents relies entirely on how accurately they perceive the environment. Prior works simplify the manufacturing environment by treating the disjunctive graph representation as a homogeneous structure. They merge the strict, directed logic of job flow with the undirected, spatial conflict of machine sharing. 

In this work, we introduced the HGT-Scheduler. We hypothesized that explicitly distinguishing between precedence constraints and contention constraints would improve the quality of the learned scheduling policy. We formulated the JSSP state strictly as a heterogeneous graph, maintaining separate adjacency matrices for \texttt{precedes} and \texttt{competes} edges. We processed this graph using a HGT, which applies type-dependent attention to learn specific logic for each distinct constraint type.

Our empirical results on the Fisher-Thompson benchmark instances strongly validate this hypothesis. On the FT06 instance, the HGT-Scheduler achieved a highly competitive 8.36\% optimality gap. Crucially, it demonstrated a statistically significant improvement over an identical architecture that merged the edge types ($p=0.011$). This controlled ablation proves that edge-type awareness directly enhances the network's ability to identify and resolve shop floor bottlenecks. Furthermore, our depth ablation study confirmed that a three-layer attention architecture provides the optimal receptive field for gathering global context without suffering from over-smoothing.

While the results are promising, we acknowledge the limitations of the current study. The performance on the larger FT10 instance revealed a clear sample-efficiency trade-off. While both transformer-based models vastly outperformed traditional heuristics and the standard GIN baseline, the heterogeneous HGT-Scheduler did not achieve a statistically significant advantage over the homogeneous Homo-HGT under our strict 50,000-step training limit. A heterogeneous graph is structurally more accurate, but it presents a wider mathematical search space. The agent requires more exploration time to calibrate the separate attention mechanisms across a larger graph.

These limitations clearly dictate the direction of our future work. First, we must extend the training horizons for larger instances like FT10 and FT20. By relaxing the strict step limit, we can allow the heterogeneous representation to fully converge and demonstrate its structural superiority at scale. Second, we plan to explore dynamic job arrivals. The current formulation addresses static scheduling instances where all jobs are known in advance. A true industrial environment is dynamic. We will investigate how the HGT-Scheduler's type-dependent attention handles the sudden insertion of new nodes and edges during an active scheduling episode. By continuing to refine how neural networks perceive complex constraint structures, we can develop increasingly robust and intelligent scheduling agents for modern manufacturing.

%Bibliography
\bibliographystyle{unsrt}  
\bibliography{paper}

\end{document}